\documentclass[11pt]{article}

\usepackage[final]{acl}

\usepackage{times}
\usepackage{latexsym}
\usepackage{comment}

\usepackage[T1]{fontenc}

\usepackage[utf8]{inputenc}

\usepackage{microtype}

\usepackage{inconsolata}

\usepackage{graphicx}

\usepackage{listings}
\usepackage{adjustbox}
\usepackage{subcaption}
\usepackage{algorithm}
\usepackage{algpseudocode}
\usepackage{amsmath} 

\title{MemToolAgent: Leveraging Memory for Tool Using Agents Based on Environment and User Feedback}

\author{
  \textbf{Suleyman Armagan Er\textsuperscript{*,1,2}},
  \textbf{Danilo Ribeiro\textsuperscript{\dag,1}},
  \textbf{Yogesh Virkar\textsuperscript{1}},
  \textbf{Surafel Lakew\textsuperscript{1}},
\\
  \textbf{Adi Kalyanpur\textsuperscript{1}},
  \textbf{James Gung\textsuperscript{1}},
  \textbf{Thomas Delteil\textsuperscript{1}},
  \textbf{Arshit Gupta\textsuperscript{1}}
\\
\\
  \textsuperscript{1}AWS AI,
  \textsuperscript{2}University of Washington
}

\begin{document}
\maketitle
\renewcommand{\thefootnote}{\fnsymbol{footnote}}
\footnotetext[2]{Corresponding Author: \href{mailto:danilrib@amazon.com}{danilrib@amazon.com}}
\footnotetext[1]{Work done during internship at AWS AI}
\begin{abstract}

Modern large language model (LLM) agents can use external tools to help users solve complex tasks. However, for problems that require learning from long-term historical events or from previous agent-environment interactions, LLM agents are required to use memory mechanisms to store and retrieve experiences. While sophisticated memory systems exist for dialogue agents, few studies have empirically examined how to improve agents' tool-using capabilities through past user-agent conversations. We propose MemToolAgent, a framework that improves tool use through memory management. Our approach contains a memory extraction module that processes past experiences into structured memory entries, and a retrieval module that dynamically selects a subset of the stored memory entries. This enables more personalized and accurate responses aligned with user preferences and feedback without requiring LLM fine-tuning. In summary, this work has three main contributions: (1) a unified memory entry format that improves both general-purpose and personalized tool use without LLM fine-tuning, (2) a reflection-based memory extraction that uses environment and user feedback to distill wrong executions into critiques to store, and (3) a retrieval module that chooses how many past experiences to use based on the memory similarity distribution. MemToolAgent achieves 29\%, 80\%, and 17\% relative improvements compared to strong baselines on the WorkBench, NESTFUL, and PEToolBench benchmarks, respectively.

\end{abstract}

\section{Introduction}
Combining powerful LLMs with additional components like memory systems, workflow mechanisms, APIs, and other tools to create LLM agents is an emerging trend. These agents are designed to solve complex tasks that require sequential reasoning and multi-step planning. In this context, tools are functions utilized by agents to interact with their environment. Integrating LLMs with external tools and APIs allows these language models to leverage various systems, enabling end-users to interact with those systems through a chat interface. An LLM agent's tool use ability, i.e., using appropriate tools for solving a task, is an important capability. 

Despite significant advancements in recent years, fixed context windows of LLMs limit maintaining coherence and thematic continuity during long real-world conversations \citep{tan2025prospect}. Due to long sequences containing abundant irrelevant information, the attention mechanism loses its effectiveness, which is not desired for critical scenarios in education and healthcare. As a result, integrating memory mechanisms into agents is crucial for long-term interaction ability. LLM agents need reliable external feedback to correct themselves \citep{kamoi2024can}, and memory can provide this feedback by storing and retrieving past interactions. Using experiences allows agents to more accurately anticipate user needs and learn from mistakes. 

\begin{figure*}[h!]
  \centering
  \includegraphics[width=0.8\linewidth]{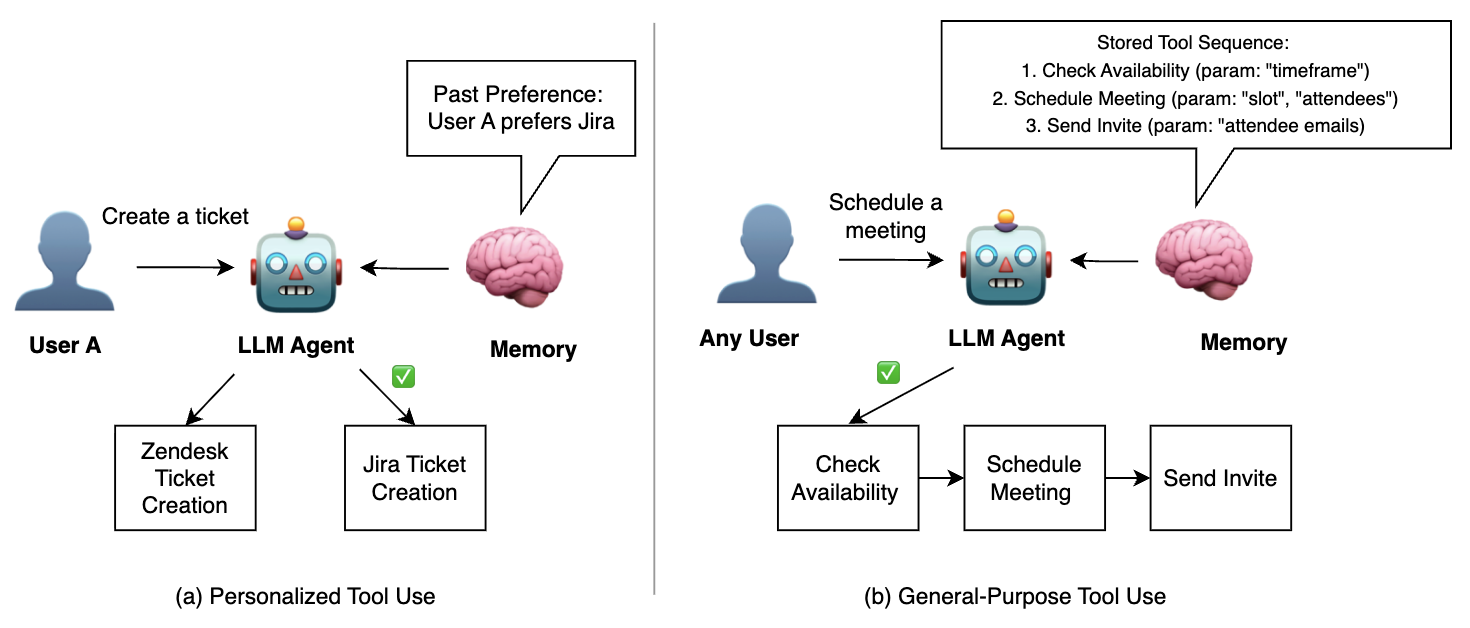}
  \caption {Integrating memory for two main tool use tasks}
  \label{fig:main_use_cases}  
\end{figure*}

Integrating memory into an LLM agent supports two main types of tasks: personalized and general-purpose tool use, as shown in Figure~\ref{fig:main_use_cases}. Users utilize the same tools in particular ways, such as using different input parameters, or they might use different tools to achieve the same goal. In scenarios where the agent has access to tools with similar functionalities, it requires user context to correctly execute the task. Storing the past interactions in the memory allows for better personalization. For instance, personalization can resolve the tool-level ambiguity if "create a ticket" could refer to Zendesk ticket creation or Jira issue creation. Regarding general-purpose tool use, there are common error patterns in tool use agents, such as dependency violation, parameter hallucination, and premature tool calling. Having past experiences in memory can help the agent use the correct parameters for tools and use the tools in the correct order. For example, this is useful in scenarios where the agent utilizes the same set of tools for a group of users to solve generic tasks like forwarding emails. 

Our work offers three main contributions: (1) a unified memory entry format of storing past experiences for better general-purpose and personalized tool use performance without LLM fine-tuning, (2) an environment and user feedback-based reflection approach for memory writing to distill failed past experiences into critiques, and (3) a retrieval module based on similarity distribution that dynamically selects relevant memory entries to guide the agent for the current task.

\section{Related Work}

There has been a growing interest in memory management for LLM agents in recent years. A related earlier idea is Case-based reasoning (CBR) \citep{aamodt1994case} where systems retrieve and use similar past cases (experiences) to solve new problems. MemGPT (MemoryGPT) \citep{packer2023memgpt} introduces an OS-inspired memory hierarchy with two memory types called the main context and the external context. More recent work includes graph-based architectures to organize the information more flexibly and expressively. In Mem0 \citep{chhikara2025mem0}, the memory is modeled as a directed knowledge graph to express complex relationships. This allows dynamic and more advanced storage and retrieval. A-MEM (Agentic Memory) \citep{xu2025mem} proposes a memory based on the Zettelkasten knowledge management system. The memory entries are linked as interconnected networks, and the existing entries are updated based on new experiences. Similarly, Zep \citep{rasmussen2025zep} and Cognitive Weave \citep{vishwakarma2025cognitive} are based on temporally aware knowledge graphs and Spatio-Temporal Resonance Graphs, respectively. 

Reflexion agents \citep{shinn2023reflexion} turn task feedback into self-reflections that are stored in an episodic memory for subsequent runs. Reflective Memory Management (RMM) \citep{tan2025prospect} uses a Reinforcement Learning based approach to rank memory entries based on their utilities in past executions. PersonaAgent \citep{zhang2025personaagent} combines episodic and semantic memory for an LLM agent with personalized memory that enables the agent to perform actions customized to the user. Agent Workflow Memory \citep{wang2024agent} stores workflows in the memory to guide the agent in web navigation tasks. EXPEREPAIR \citep{mu2025experepair} introduces an agent with a dual-memory mechanism for code repair. T1 \citep{chakraborty2025t1} uses a simple cache for agents to store tool call results. However, the main goal is to reuse the previously accessed data. 

\begin{figure*}[h!]
  \centering
  \includegraphics[width=0.9\linewidth]{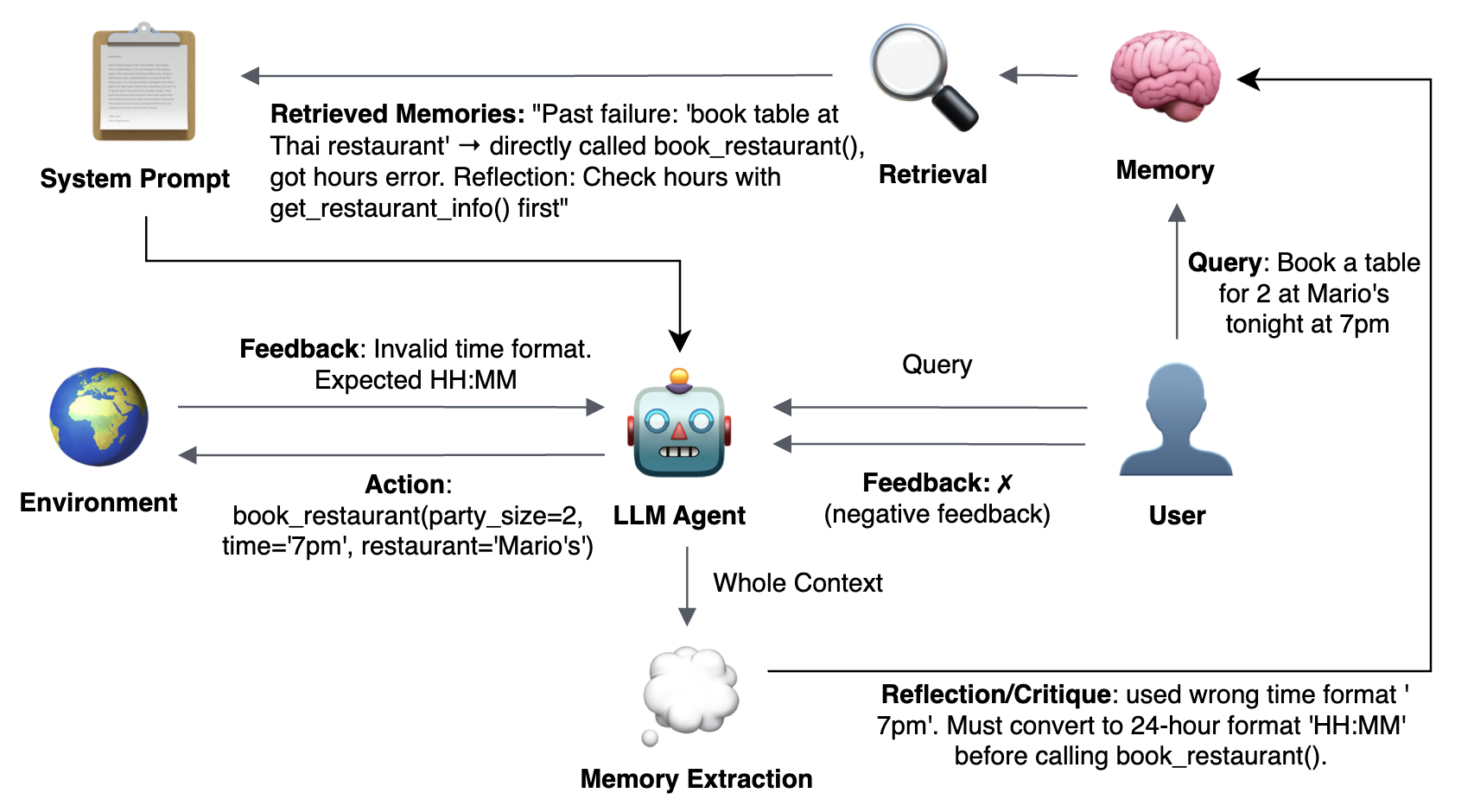}
  \caption {MemToolAgent overview with a simple restaurant booking scenario where the agent retrieves similar memories, receives feedback on an invalid time format, and generates a reflection to update its memory}
  \label{fig:overall}  
\end{figure*}

$Mem^p$ \citep{fang2025memp} is a procedural memory framework that extracts both fine-grained instructions and high-level abstractions from past agent trajectories. Memento \citep{zhou2025memento} stores past experiences in an episodic memory, learns and continually updates a case-selection policy based on environmental feedback. Training-Free GRPO \citep{cai2025training} is a training-free alternative to agentic RL that optimizes an experience library consisting of natural-language experiential knowledge. TOOLMEM \citep{xiao2025toolmem} stores memories of tool capabilities from previous interactions and helps the agent select the most suitable tool for a given task at inference. LEGOMem \citep{han2025legomem} is a procedural memory approach for multi-agent systems that distills only successful task trajectories into memory entries with brief reflections. ReMem agents \citep{wei2025evo} combine reasoning, action, and memory refinement and actively exploit, prune, and reorganize their memory while solving the current task. 

There are in-context learning (ICL) approaches closely related to agents with memory. For example, STE \citep{wang2024llms} creates tool use trajectories based on a trial-and-error approach offline and stores them in a static memory for inference. In-Context Abstraction Learning (ICAL) \citep{sarch2024vlm} creates memory abstractions from sub-optimal trajectories based on human-in-the-loop learning and self-reflection. 

Unlike the related work, MemToolAgent addresses both personalized and general-purpose tool use. It proposes a general memory entry format and stores and leverages detailed information like tool trajectories and a reflection/critique about the execution. Importantly, MemToolAgent also incorporates user feedback as a signal to improve the quality of memory entries.

\section{Method: MemToolAgent}

MemToolAgent consists of two main components: the memory retrieval and memory extraction modules. Figure~\ref{fig:overall} illustrates the general overview of our proposed approach. Memory entries, i.e., past experiences, are retrieved from the memory based on the semantic similarity between the query and the entries. Dynamic "n" Top-n Embedding-based Retrieval is proposed for retrieval. Instead of retrieving a fixed number of memory entries for each query, the number of entries to retrieve is dynamically determined based on the cosine similarities between the embeddings of memory entries’ queries and the test query. Then, the entries are used as in-context learning to guide the agent for the current task. 

To add a new entry to the memory based on the current query, an LLM-based "Memory Extraction" module provides a reflection/critique of the execution, which is then stored in the memory with the query and user feedback. 

MemToolAgent specifies what to store in memory with a general memory entry format. Each memory entry $m$ is represented as a tuple $m = (q, a, f, r)$ where:
\begin{itemize}
    \item $q$: the user query
    \item $a = \{a_1, a_2, \ldots, a_n\}$: the sequence of tool calls executed by the agent
    \item $f \in \{0, 1\}$: binary user feedback indicating task success ($f=1$) or failure ($f=0$)
    \item $r$: reflection or critique generated by the memory extractor when $f=0$ (otherwise $r = \emptyset$)
\end{itemize}

 With binary user feedback, we simulate the scenarios where the user does not know anything about the underlying tools/APIs and gives feedback based on the outcome. However, this feedback could easily be replaced by any other type of feedback, such as text or score. Figure~\ref{fig:workbench_entry_with_negative_feedback} illustrates an example memory entry with reflection.
 
\begin{figure*}[t]
\begin{lstlisting}[basicstyle=\ttfamily\small, frame=single, escapeinside={(*}{*)}]
(*\textbf{Query:}*) Was total visits more than 21 at any time in the last 1 weeks? If so, please
plot it as a line chart
(*\textbf{Answer (Tool Calls):}*) ['analytics.total_visits_count.func(time_min="2023-11-23",
time_max="2023-11-30")', 'analytics.create_plot.func(time_min="2023-11-23",
time_max="2023-11-30", value_to_plot="total_visits", plot_type="line")']
(*\textbf{User Feedback:}*) Incorrect
(*\textbf{Reflection:}*) The agent created the line chart despite the condition not being met. The 
agent concluded correctly that the total visits did not exceed 21 at any point
(the highest was 19 visits on 2023-11-28), but then proceeded to create the
line chart anyway. This contradicts the original instruction which specified to plot
the data ONLY IF visits exceeded 21.
\end{lstlisting}
\caption{Memory entry with negative feedback for WorkBench}
\label{fig:workbench_entry_with_negative_feedback}
\end{figure*}

\subsection{Memory Retrieval: Dynamic "n" Top-n Embedding-based Retrieval}

Retrieving a fixed number of memory entries for every query is not optimal since some queries might not have relevant memory entries, while there might be many similar entries for other queries. Prior work \citep{taguchi2025efficient, xu2025cluster} in retrieval-augmented generation (RAG) addresses the trade-off between missing information with too small k and adding redundancy with too large k, where k is the number of retrieved documents. These methods dynamically decide the number k and show that the optimal amount of retrieved context varies across queries. Adaptive-k \citep{taguchi2025efficient} selects "k" as the largest adjacent similarity-score gap.  

Our Dynamic "n" approach has four main steps illustrated by the Algorithm~\ref{alg:dynamic_n_symbols}. The first two steps are calculating the cosine similarity values between the query and memory entries and sorting them in descending order. The third step calculates a first-order derivative estimate of sorted similarities based on a sliding window approach where \(x\) is an array of sorted similarities, and $R$ is considered as the "radius" of the derivative estimate. The final step is finding the peaks, i.e., sudden drops in similarity, ideally showing a class boundary. The peak corresponds to our “n” value for Dynamic "n" retrieval. The entire process is inspired by the concept of finding “knees” for designing systems \citep{satopaa2011finding}. The \texttt{find\_peaks}\footnote{\url{https://docs.scipy.org/doc/scipy/reference/generated/scipy.signal.find_peaks.html}} function from the SciPy library is utilized for peak finding. 

\begin{figure}[h!]
  \includegraphics[width=0.99\linewidth]{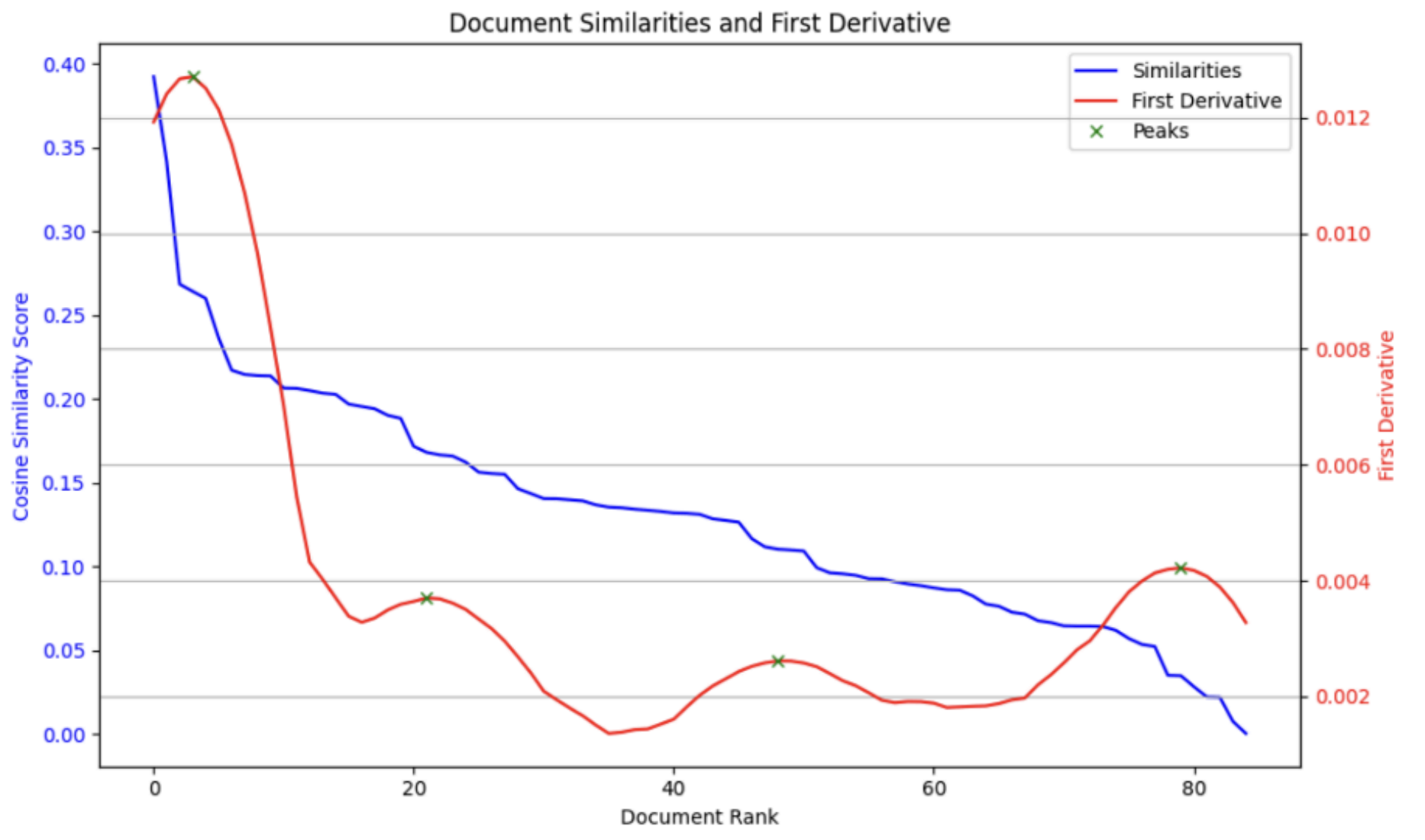}
  \caption {Descending similarity values (blue), derivative estimate (red), and the peaks}
  \label{fig:dynamic_n}  
\end{figure}

Figure~\ref{fig:dynamic_n} illustrates an example. The blue line shows the sorted similarity values, while the red line shows the first derivative estimation. As observed, the locations where there are local maxima, i.e., a peak, correspond to locations with a sudden drop in similarity. Dynamic "n" retrieval has three parameters, namely R, prominence, and the peak number. R is a smoothing parameter, and larger R values result in smoother derivative estimates. Prominence is a parameter of the \texttt{find\_peaks} function to filter important peaks based on comparison of neighboring values. The peak number is simply which peak to select.

Similarity curves can be noisy, have multiple knees and multi-step drops rather than huge gaps. We hypothesize that finding peaks based on the derivative estimate rather than raw similarity values results in a more robust approach than Adaptive-k. Dynamic "n" robustly finds points of change or cluster boundaries on noisy similarity curves, while Adaptive-k finds the biggest drop and works well only when the curve has one clean knee. 

\begin{algorithm}[H]
\caption{Dynamic $"n"$ Retrieval}
\label{alg:dynamic_n_symbols}
\begin{algorithmic}[1]
\State \textbf{input:} $q$ (query), $\mathcal{M}$ (memory), $R$ (radius), peak number, prominence
\State \textbf{output:} $n$ (number of entries to select)
\State $S \gets [\text{CalculateSimilarity}(q, m)$
\Statex $\qquad \text{ for } m \in \mathcal{M}]$
\State $x \gets \text{SortDescending}(S)$ 

\For{$j = R$ \textbf{to} $|x| - R - 1$}
    \State $y(j) \gets \left[ -\frac{\sum_{\tau=-R}^{R} \tau x(j + \tau)}{\sum_{\tau=-R}^{R} \tau^2} \right]$
\EndFor

\State $peaks \gets \text{\(find\_peaks\)}(y, prominence)$ 
\State $n \gets peaks[peak\_number]$ 
\end{algorithmic}
\end{algorithm}

\subsection{Memory Extraction Module}
The agent needs to apply experience to the current task based on the user feedback and the context. An important observation from our experiments is that agents struggle to effectively learn from retrieved negative experiences when they are presented as raw execution traces. We hypothesize this is because the agent must simultaneously understand what went wrong in the past example while solving the current task. To make the agent’s work easier, a "Memory Extraction" module is proposed. It reflects on the context and user feedback to understand what was wrong. For wrong executions, the reflection/critique is added to the memory instead of the raw context, which might be too complicated and contain irrelevant parts. Doing the reflection as a separate step helps the agent focus only on the current task and learn from its mistakes.

The memory extraction module is an LLM $\mathcal{E}$ that takes the following as input: (1) current query $q$, (2) set of available tools $\mathcal{T} = \{t_1, t_2, \ldots, t_m\}$, (3) agent's response (actions) $a = \{a_1, a_2, \ldots, a_n\}$ including tool calls, (4) environment feedback $e$ including tool responses, (5) user feedback $f \in \{0, 1\}$, and (6) retrieved semantically similar memory entries $\mathcal{M}' \subset \mathcal{M}$. The module outputs a reflection/critique $r = \mathcal{E}(q, \mathcal{T}, a, e, f, \mathcal{M}')$.

Using retrieved memory entries that are semantically similar to the current user query is important. The main use case for general-purpose tool use is having different queries with common mistakes, which are only sometimes detected by the memory extractor. A similar entry with a correct reflection could help the memory extractor produce a correct critique. For personalized tool use, memory extractor is slightly different. Semantically similar memory entries are used to find user patterns and preferences. The memory extraction module is used to extract high-level user preferences, such as “User prefers tools under \(API\_X\)". 

\section{Evaluation}
Memory can be applied to multiple problem settings in tool-using agents. Our experiments focus on two main tasks: (1) general-purpose tool use and (2) personalization. Three benchmarks, WorkBench \citep{styles2024workbench}, NESTFUL \citep{basu2024nestful}, and PEToolBench \citep{xu2025petoolllm}, were used for evaluation. WorkBench and NESTFUL are general-purpose tool use benchmarks, while PEToolBench is a personalized tool orchestration benchmark. 

\begin{table*}[h!]
\centering
\small
\renewcommand{\arraystretch}{1.2}
\setlength{\tabcolsep}{4pt}  
\begin{adjustbox}{max width=\textwidth}
\begin{tabular}{l|cccccc|ccccc}
\hline
\multicolumn{1}{c|}{} & \multicolumn{6}{c|}{\textbf{WorkBench Accuracy}} & \multicolumn{5}{c}{\textbf{NESTFUL}} \\
\hline
\textbf{Method} & \textbf{Analytics} & \textbf{Calendar} & \textbf{CRM} & \textbf{Email} & \textbf{Multi-Domain} & \textbf{Overall} & \textbf{F1 Func.} & \textbf{F1 Param.} & \textbf{Part. Acc.} & \textbf{Full Acc.} & \textbf{Win Rate}\\
\hline
No Memory & 18.52 & 73.74 & 69.44 & 79.01 & 57.67 & 57.56 & \textbf{96.30} & 61.30 & 31.90 & 15.60 & 51.20 \\
Extended Thinking & 38.89 & 85.86 & 75 & 86.42 & 53.44 & 64.12 & 95.20 & 62.70 & 33.40 & 16.90 & 66.10 \\
ReAct Agent & 27.78 & 91.92 & 72.22 &  90.12 & 61.38 & 65.94 & 93.20 & 56.40 & 29.50 & 10.80 & 58.20\\
MemToolAgent & \textbf{86.11} & \textbf{92.93} & \textbf{88.89} & \textbf{97.53} & \textbf{73.54} & \textbf{85.06} & 96.00 & \textbf{78.00} & \textbf{40.60} & \textbf{30.40} & \textbf{70.20} \\
\hline
\end{tabular}
\end{adjustbox}
\caption{MemToolAgent Results for General Purpose Tool Use Benchmarks}
\label{tab:main_general_purpose_results}
\end{table*}

\subsection{Benchmarks}
WorkBench is an outcome-centric tool use benchmark that evaluates agents on workplace tasks such as sending emails and scheduling meetings. It includes five databases and 27 tools. WorkBench tasks are from different domains like Analytics, Calendar, etc.  The evaluation metric is accuracy, which is defined as the percentage of tasks where the outcome from the agent’s actions matches the expected outcome. The correct outcome means the correct changes in the database values.

WorkBench queries were created based on pre-defined templates such as 'Delete an email from person X' or 'Schedule a meeting with person Y'. There are 10 queries per template. Since the benchmark has no train-test split, one query from each template is used to populate the memory. This ensures that there is at least one directly relevant memory entry for each test query. This setup is used to show the advantages of using the memory. The test set consists of 108 Analytics, 99 Calendar, 72 CRM, 81 Email, and 189 Multi-domain tasks.

NESTFUL benchmark evaluates LLMs on nested sequences of tool calls, where the output of one tool call is used as input for the next tool call. It includes 1891 test queries based on MathQA \citep{amini2019mathqa} and StarCoder2-Instruct \citep{wei2024selfcodealign} benchmarks. The first 500 tasks are used as test set while the next 500 tasks are used to populate the memory. The benchmark uses five evaluation metrics, which are F1 scores for tool and parameter generation, Partial Sequence Matching, Full Sequence Matching, and Win Rate. Partial Sequence Matching measures how many predicted APIs with their argument-value pairs match the gold API sequence, Full Sequence Matching evaluates whether the model produces the exact full sequence, and Win Rate assesses whether the predicted APIs lead to the gold answer when executed.

PEToolBench is a personalized tool orchestration benchmark. Each query comes with an instruction history consisting of 80-100 past user queries, where each past query includes a binary rating (1 for correct/desired execution, 0 otherwise). For our experiments, this instruction history forms the memory, where each memory entry consists of a past query, its corresponding tool call, and binary user feedback. Test queries are designed such that the agent must rely on user preferences from memory for correct tool execution. A subset of 591 queries is used as the test set after filtering the queries with sufficiently long instruction histories (at least 80 past queries). This makes sure that the agent needs a memory rather than using the full history in context. The benchmark uses two evaluation metrics: Tool Accuracy, which evaluates the model's ability to call the appropriate tool based on user preferences, and Parameter Accuracy, which assesses whether the model generates the correct parameters.

\subsection{Experiment Results}
This section presents the main experiments with details, ablation studies demonstrating the contribution of individual MemToolAgent parts, and a cost analysis.

\subsubsection{Experiment Settings}
Claude 3.5 Sonnet \citep{anthropic2024claude35sonnet} is used for PEToolBench results while NESTFUL and WorkBench experiments use Claude 3.7 Sonnet \citep{anthropic2025claude37sonnet}. Qwen3-Embedding-8B \citep{zhang2025qwen3} embedding model is used for memory retrieval. All experimental results are based on a single run.

Dynamic "n" retrieval hyperparameters are manually tuned. R is set to 10 and prominence is set to $10^{-5}$. Empirically, using the second peak for PEToolBench performs better while the first peak is the optimal choice for WorkBench and NESTFUL.

\subsubsection{General Purpose Tool Use Experiments}

Table~\ref{tab:main_general_purpose_results} compares MemToolAgent with no-memory, extended thinking, and ReAct \citep{yao2022react} in NESTFUL and WorkBench. For NESTFUL, MemToolAgent nearly doubles the Full Sequence Matching performance (from 15.6\% to 30.4\%) compared to the no-memory baseline, while the Win Rate increases from 51\% to 70\%. Enabling extended thinking improves Win Rate but yields only a slight increase in Full Sequence Matching. These results demonstrate that extra reasoning alone is not enough, and the model requires concrete examples from memory to properly follow the tool calling format. MemToolAgent demonstrates significant performance improvements on WorkBench, with Analytics task accuracy increasing from 18\% to 86\% and overall accuracy rising from 57\% to 85\% compared to the no-memory baseline.

\subsubsection{Personalized Tool Orchestration Experiments}

\begin{table}[!t]
\centering
\small
\renewcommand{\arraystretch}{1.2}
\setlength{\tabcolsep}{4pt}  
\begin{tabular}{lcc}
\hline
\textbf{Method} & \textbf{Tool Accuracy} & \textbf{Parameter Accuracy}\\
\hline
A-MEM & 0.700 & 0.820 \\
Mem0 & 0.700 & 0.820  \\
MemToolAgent & \textbf{0.820} & 0.820 \\
\hline
\end{tabular}
\caption{MemToolAgent Results for PEToolBench}
\label{tab:petoolbench_main_results}
\end{table}

Table~\ref{tab:petoolbench_main_results} shows MemToolAgent's performance compared to state-of-the-art memory architectures A-MEM and Mem0 on PEToolBench. Due to the cost of LLM calls for memory addition in these methods and the MemToolAgent with memory extractor, the experiments used a 50-query subset consisting of the first 50 queries from the 591-query test set. Top-15 embedding-based retrieval was used for A-MEM and Mem0. MemToolAgent achieves a Tool Accuracy of 0.82, representing a 17\% relative improvement over A-MEM and Mem0. This demonstrates that MemToolAgent is more effective at leveraging memories for selecting the appropriate tool based on user preferences. All methods perform equally well at generating the appropriate parameters.

\begin{figure*}[htb]
  \centering
  \includegraphics[width=0.8\linewidth]{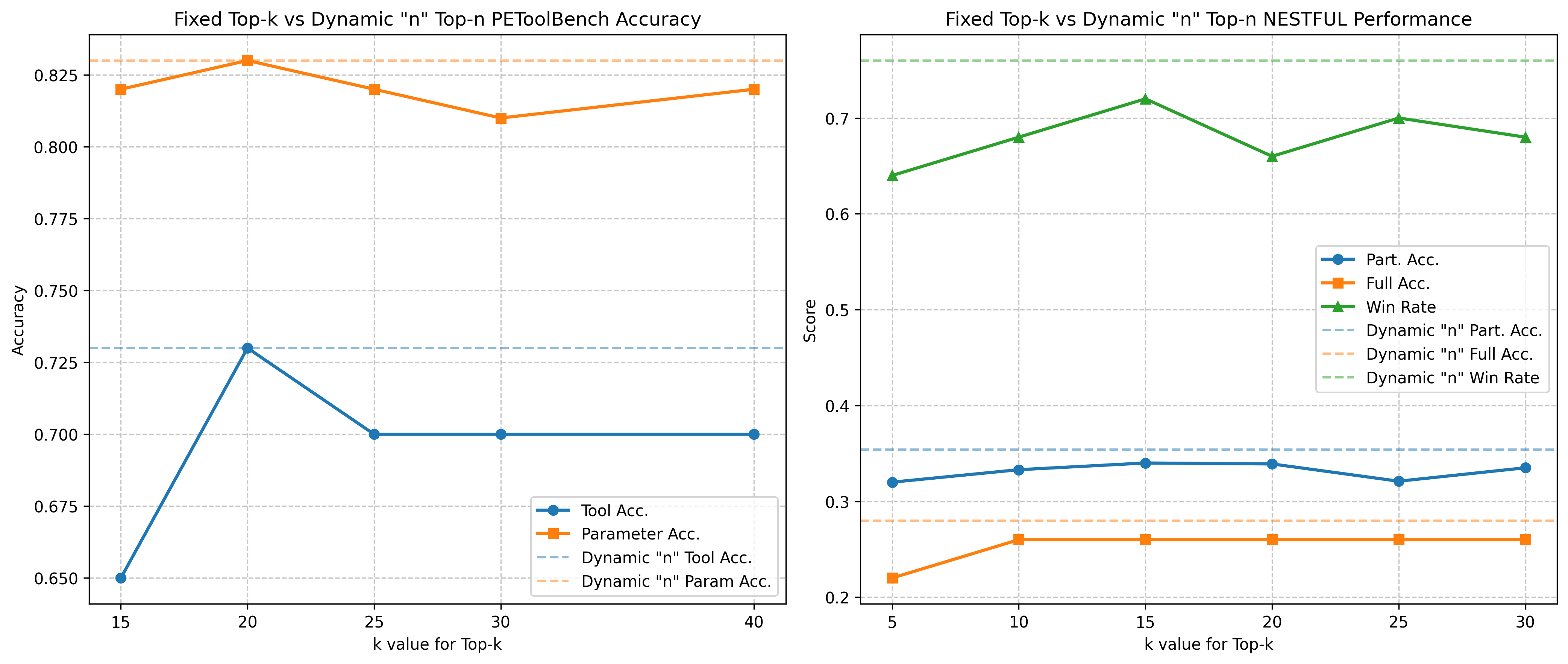}
  \caption {Ablation studies comparing fixed top-k memory retrieval with the Dynamic top-n approach}
  \label{fig:topk_vs_dynamic_n}  
\end{figure*}

\begin{table*}[h!]
\centering
\small
\renewcommand{\arraystretch}{1.2}

\begin{subtable}{\textwidth}
\centering
\setlength{\tabcolsep}{4pt}
\begin{adjustbox}{max width=\textwidth}
\begin{tabular}{l|cc|ccccc}
\hline
\multicolumn{1}{c|}{} & \multicolumn{2}{c|}{\textbf{PEToolBench}} & \multicolumn{5}{c}{\textbf{NESTFUL}} \\
\hline
\textbf{Method} & \textbf{Tool Acc.} & \textbf{Param Acc.} & \textbf{F1 Func.} & \textbf{F1 Param.} & \textbf{Part. Acc.} & \textbf{Full Acc.} & \textbf{Win Rate}\\
\hline
MemToolAgent w/o memory extractor  & 0.700 & 0.800 & 0.930 & 0.619 & 0.327 & 0.160 & 0.548\\
MemToolAgent & 0.820 & 0.820 & 0.960 & 0.755 & 0.397 & 0.290 & 0.666\\
\hline
\end{tabular}
\end{adjustbox}
\caption{Effect of memory extraction module on MemToolAgent Performance}
\label{tab:thinker_ablation_studies}
\end{subtable}

\vspace{0.3cm}

\begin{subtable}{\textwidth}
\centering
\begin{adjustbox}{max width=\textwidth}
\begin{tabular}{lcccccc}
\hline
\textbf{Method} & \textbf{Analytics} & \textbf{Calendar} & \textbf{CRM} & \textbf{Email} & \textbf{Multi-Domain} & \textbf{Overall}\\
\hline
MemToolAgent w/o memory extractor & 38.89 & 88.89 & 77.78 & 91.36 & 65.08 & 69.76 \\
MemToolAgent w/ paraphrased queries and w/o memory extractor & 37.96 & \textbf{97.98} & 76.39 & 85.19 & \textbf{70.37} & 71.95 \\
MemToolAgent w/ synthetic queries and w/o memory extractor & 28.70 & 85.86 & 76.39 & 91.36 & 54.50 & 63.39 \\
MemToolAgent & \textbf{47.22} & 93.94 & \textbf{87.50} & \textbf{92.59} & 67.20 & \textbf{74.50} \\
\hline
\end{tabular}
\end{adjustbox}
\caption{Comparison of Different Memory Types for WorkBench}
\label{tab:workbench_memory_type_ablation}
\end{subtable}

\caption{Ablation Studies for memory extraction module and Memory Types}
\label{tab:ablation_studies}
\end{table*}

\subsection{Ablation Studies}

\subsubsection{Comparing Fixed Top-k Retrieval with Dynamic Top-n}

Figure~\ref{fig:topk_vs_dynamic_n} compares fixed top-k with the Dynamic top-n memory approach on 100 PEToolBench queries (the first 100 queries from the 591-query test set). Dynamic top-n retrieval finds a sweet spot for the optimal number of memory entries. The figure also demonstrates that Dynamic top-n outperforms the fixed top-k method across different k values on the first 50 queries from the NESTFUL test set. The results confirm our hypothesis that the optimal number of memory entries depends on the test query, and the proposed Dynamic top-n approach moves us in that direction.

\subsubsection{Effect of Memory Extraction module}

Tables~\ref{tab:thinker_ablation_studies} and \ref{tab:workbench_memory_type_ablation} show the performance improvements achieved by incorporating the memory extraction module into MemToolAgent. Due to cost and time constraints, PEToolBench uses 50 queries while NESTFUL and WorkBench use full test sets. Using the memory extractor to store high-level user preferences in memory improves Tool Accuracy in PEToolBench. Furthermore, NESTFUL experiments with top-5 retrieval show that adding reflection/critique to memory significantly boosts performance. WorkBench experiments with top-1 retrieval further demonstrate that reflections enable better learning from mistakes than raw execution context alone.

\subsubsection{Hard Preference Guiding and Simple Ranking for PEToolBench}

Table~\ref{tab:petoolbench_hard_guide_ranker_ablation} demonstrates the effectiveness of simple ranker and Hard Preference Guiding ideas through ablation studies. The original approach for PEToolBench provides the agent with the entire instruction history, which is not optimal, as irrelevant queries can make the task more challenging. Embedding-based Retrieval\_v1 improves this by retrieving only the top-10 semantically similar queries from memory. Retrieval\_v2 builds on v1 by introducing a simple ranker that separates memory entries by rating (1 vs. 0) in the system prompt to emphasize user preferences. This is a small modification for personalization tasks. Although it substantially increases the tool accuracy, there is a decrease in parameter accuracy. This is due to binary feedback, which only provides correct/incorrect information. Negative feedback causes the agent to consider correct parameters as incorrect, as the tool itself can be incorrect even with correct parameters.

Error analysis revealed cases where the agent selected tools despite negative user preferences from the memory. To address this, we introduce an alternative baseline called Hard Preference Guiding which removes tools with a rating of 0 from the system prompt. This is a substitute for the memory extraction module's approach of providing user preferences as soft guidance. It provides a cost-efficient alternative to the LLM-based memory extraction module, eliminating the need for additional LLM calls during memory addition. Hard Preference Guiding extends MemToolAgent w/o memory extractor and improves Tool Accuracy from 0.72 to 0.84, as shown in Table~\ref{tab:petoolbench_hard_guide_ranker_ablation}. Although less general than memory extraction, Hard Preference Guiding is effective for scenarios like PEToolBench and could be used as a complementary approach.

\begin{table}[!t]
\centering
\small
\renewcommand{\arraystretch}{1.2}
\setlength{\tabcolsep}{2pt}  
\begin{adjustbox}{max width=\textwidth}
\begin{tabular}{lcc}
\hline
\textbf{Method} & \textbf{Tool Acc.} & \textbf{Param Acc.}\\
\hline
Original Approach & 0.509 & \textbf{0.864} \\
Embedding-based Retrieval\_v1 & 0.538 & 0.851 \\
Embedding-based Retrieval\_v2 & 0.653 & 0.817 \\
MemToolAgent w/o memory ext. & 0.722 & 0.849 \\
Hard Preference Guiding  & \textbf{0.844} & 0.827 \\
\hline
\end{tabular}
\end{adjustbox}
\caption{MemToolAgent Ablation Results for PEToolBench}
\label{tab:petoolbench_hard_guide_ranker_ablation}
\end{table}

\subsubsection{Experiments with Different Memory Entry Types}

Table~\ref{tab:workbench_memory_type_ablation} compares different memory entry formats for WorkBench. The "MemToolAgent w/ paraphrased queries and w/o memory extractor" setting uses paraphrased versions of queries that are used for generating memory entries while ensuring they have the same ground truth (GT). Everything else remains identical to MemToolAgent w/o memory extractor. Since the memory entries and test queries are similar to each other, this setting simulates a scenario where memory entries are slightly different. Using different query formulations can expand the agent's exploration space. Using paraphrased queries leads to better performance on Multi-domain and Calendar tasks than original queries. 

The "MemToolAgent w/ synthetic queries and w/o memory extractor" setting addresses the cold-start scenario where no queries exist to populate the memory before inference. Given the tool descriptions, an LLM generates synthetic queries, which are then executed and added to the memory. Using synthetic queries improves results compared to the no-memory baseline, except on multi-domain tasks. For multi-domain tasks, it is challenging to generate queries that are both complex enough and sufficiently similar to the test set. Nevertheless, this still demonstrates that synthetic queries can help avoid starting with an empty memory.

\subsection{Token Usage Analysis}

Table~\ref{tab:cost_analysis} shows the total number of input and output tokens used by Claude 3.7 Sonnet for 50 NESTFUL test queries. Although MemToolAgent increases the total number of input tokens, the number of output tokens has a much lower increase compared to the Extended Thinking and ReAct baselines. This is an advantage given that output tokens tend to be more expensive and contribute more to the overall response latency.

\begin{table}[!t]
\centering
\small
\renewcommand{\arraystretch}{1.2}
\setlength{\tabcolsep}{4pt}  
\begin{tabular}{lcc}
\hline
\textbf{Method} & \textbf{Input Tokens} & \textbf{Output Tokens} \\
\hline
No Memory & \textbf{72345} & \textbf{10121} \\
Extended Thinking  & 73795 & 40765 \\
ReAct & 85245 & 22923 \\
MemToolAgent & 307822 & 14585 \\
\hline
\end{tabular}
\caption{Total number of input and output tokens}
\label{tab:cost_analysis}
\end{table}

\section{Conclusion}

MemToolAgent is a memory framework that enhances the tool-using capabilities of LLM agents by using past experiences. The results show significant improvements across multiple benchmarks in both personalized and general-purpose tool use scenarios. The main contributions of the framework include (1) a retrieval mechanism that dynamically determines the number of relevant memory entries to retrieve, (2) a memory extraction module that generates reflections on incorrect executions to help avoid similar mistakes, and (3) an overall effective approach for both general-purpose and personalized tool use based on user and environment feedback.

\section*{Limitations}

Despite the improvements in general-purpose and personalized tool use performance achieved by MemToolAgent, several limitations remain for future work. One limitation is the cold-start problem of memory. Moreover, the benchmarks used contain fewer than 40 tools. However, there are typically more tools in real-life applications. As the number of tools increases, it is more challenging to cover the usage of all tools with memory. This could potentially make the cold-start problem more severe. Our experiments using synthetically generated queries show that even a simple approach can help the cold-start problem.

Another limitation emerges in conversational scenarios with multiple turns. The benchmarks used in our experiments are single-turn. For single-turn tasks, the memory retrieval is done once at the beginning of the conversation. This might not be optimal for multiple-user turn tasks where the agent needs to obtain the necessary information from the user across multiple turns. One possible solution is to extend MemToolAgent to retrieve from memory after each user turn, so that the agent can focus on the sub-task associated with that specific turn. 

\section*{Ethical Considerations}

The proposed MemToolAgent approach has been implemented and evaluated using the publicly available NESTFUL, WorkBench, and PEToolBench benchmarks.


\bibliography{custom}

\appendix

\section{Appendix}
\label{sec:appendix}

\begin{table*}[h!]
\centering
\small
\renewcommand{\arraystretch}{1.2}
\setlength{\tabcolsep}{3pt}  
\begin{adjustbox}{max width=\textwidth}
\begin{tabular}{lcccccc}
\hline
\textbf{Method} & \textbf{Analytics} & \textbf{Calendar} & \textbf{CRM} & \textbf{Email} & \textbf{Multi-Domain} & \textbf{Overall}\\
\hline
No Memory & 50 & 91.92 & 68.06 & 87.65 & 57.14 & 67.94 \\
Extended Thinking & 55.56 & 88.89 & 76.39 & \textbf{88.89} & 57.14 & 69.76 \\
ReAct & 48.15 & 90.91 & 79.17 & \textbf{88.89} & 58.20 & 69.40 \\
MemToolAgent & \textbf{92.59} & \textbf{93.94} & \textbf{84.72} & \textbf{88.89} & \textbf{68.78} & \textbf{83.06} \\
\hline
\end{tabular}
\end{adjustbox}
\caption{WorkBench Accuracy Results with Claude 4 Sonnet}
\label{tab:workbench_claude_4_results}
\end{table*}

\begin{figure*}[t]
\begin{lstlisting}[basicstyle=\ttfamily\small, frame=single, escapeinside={(*}{*)}]
You are an agent responsible for analyzing the INCORRECT execution of a tool-use
agent.

Today's date is Thursday, 2023-11-30 and the current time is 00:00:00.
Remember the current date and time when answering queries. Meetings must not start
before 9am or end after 6pm.

The agent had access to these tools:
{TOOL DESCRIPTIONS}

Here are similar queries, their executions, and user feedback:
{SIMILAR ENTRIES FROM MEMORY}

For each query where user feedback indicates incorrect execution, reflect on what
went wrong and use these insights.

The whole agent execution including tool calls and corresponding responses is:
{AGENT EXECUTION}

Check if the agent used a correct order of tools
Check if the agent hallucinated any tool parameters
Check if the agent used wrong values for tool parameters
Check if the agent only used the information from tools or directly from user.
Check if the agent had a correct temporal and conditional reasoning.

Before sending any error message, verify you're not contradicting yourself

Remember this is a wrong response, there is at least one mistake in tool usage

Reflect on the query, response, and other similar queries to catch the mistake,
try to keep your response short.

\end{lstlisting}
\caption{Memory extraction module's system prompt for general purpose tool use (WorkBench)}
\label{tab:thinker_system_prompt_general}
\end{figure*}

\begin{figure*}[t]
\begin{lstlisting}[basicstyle=\ttfamily\small, frame=single, escapeinside={(*}{*)}]
(*\textbf{Query:}*) We're moving Dakota Wilson to Raj. Can you make that change in the crm?
(*\textbf{Answer:}*) ['search_customers.func(customer_name="Dakota Wilson")',
'find_email_address.func(name="Raj")',
'update_customer.func(customer_id="00000076", field="assigned_to_email",
new_value="raj.patel@atlas.com")']
(*\textbf{User Feedback:}*)
Execution: Correct
\end{lstlisting}
\caption{Memory entry with positive feedback for WorkBench}
\label{tab:workbench_entry_with_positive_feedback}
\end{figure*}

\begin{figure*}[t]
\begin{lstlisting}[basicstyle=\ttfamily\small, frame=single, escapeinside={(*}{*)}]
(*\textbf{Test query:}*) 
I haven't spoken to Quinn Thomas in a while. Can you check if it's been over
14 days? If so, book a 30-minute meeting with whoever is assigned to them called
'Update on Quinn Thomas' at the first time I'm free tomorrow

(*\textbf{Related memory entry:}*)
Query: If we haven't spoke to Cameron Anderson in the past fortnight book a
30-minute meeting with whoever is assigned to them called
'Update on Cameron Anderson' at the first time I'm free tomorrow

Answer: ['customer_relationship_manager.search_customers.func(customer_name=
"Cameron Anderson")', 'calendar.search_events.func(time_min="2023-12-01 09:00:00",
time_max="2023-12-01 18:00:00")', 'calendar.create_event.func(event_name="Update on
Cameron Anderson", participant_email="sofia.santos@atlas.com",
event_start="2023-12-01 13:00:00", duration="30")']

User Feedback:
Execution: Correct

(*\textbf{Incorrect response of the agent without memory:}*)
['customer_relationship_manager.search_customers.func(customer_name="Quinn Thomas")'
,'company_directory.find_email_address.func(name="Akira Sato")',
'company_directory.find_email_address.func(name="Akira")',
'calendar.create_event.func(event_name="Update on Quinn Thomas",
participant_email="akira.sato@atlas.com", event_start="2023-12-01 09:00:00",
duration="30")']

(*\textbf{Correct response of the agent with memory:}*)
['customer_relationship_manager.search_customers.func(customer_name="Quinn Thomas")'
,'calendar.search_events.func(time_min="2023-12-01 09:00:00", time_max="2023-12-01
18:00:00")', 'calendar.create_event.func(event_name="Update on Quinn Thomas",
participant_email="akira.sato@atlas.com", event_start="2023-12-01 13:00:00",
duration="30")']

\end{lstlisting}
\caption{Example from WorkBench where a positive memory entry helps}
\label{tab:workbench_positive_entry_helps}
\end{figure*}

\begin{figure*}[t]
\begin{lstlisting}[basicstyle=\ttfamily\small, frame=single, escapeinside={(*}{*)}]
(*\textbf{Test query:}*) Can you make a line chart of total visits since September 24 if it was 
more than 21 since at any time September 24?

(*\textbf{Related memory entry:}*)
Query: was total visits less than 3 at any time since September 13? If so,
please plot it as a line chart

Answer: ['analytics.total_visits_count.func(time_min="2023-09-13",
time_max="2023-11-30")', 'analytics.create_plot.func(time_min="2023-09-13",
time_max="2023-11-30", value_to_plot="total_visits", plot_type="line")']

User Feedback: Incorrect
Reflection: The primary mistake was a logical error in interpreting the data and 
understanding the user's question. The agent correctly retrieved the data showing no 
days had less than 3 visits but incorrectly reported October 1 as having exactly 3 
visits (which is correct but irrelevant). It generated the plot as requested despite 
the condition "If so" not being met (no days had less than 3 visits). The agent 
should have clearly stated that the condition was not met.

(*\textbf{Incorrect response of the agent without memory:}*)
['analytics.total_visits_count.func(time_min="2023-09-24",
time_max="2023-11-30")', 'analytics.create_plot.func(time_min="2023-09-24",
time_max="2023-11-30", value_to_plot="total_visits", plot_type="line")']

(*\textbf{Correct response of the agent with memory:}*)
['analytics.total_visits_count.func(time_min="2023-09-24",time_max="2023-11-30")']

\end{lstlisting}
\caption{Example from WorkBench benchmark where a negative memory entry helps}
\label{tab:workbench_negative_entry_helps}
\end{figure*}

\begin{figure*}[t]
\begin{lstlisting}[basicstyle=\ttfamily\small, frame=single, escapeinside={(*}{*)}]
Today's date is Thursday, 2023-11-30 and the current time is 00:00:00. Remember the
current date and time when answering queries. Meetings must not start before 9am or
end after 6pm.

Here are similar queries, their executions, and user feedback:
{RETRIEVED MEMORIES}

For each query where user feedback indicates incorrect execution, reflect on what
went wrong and use these insights to avoid similar mistakes. Consider these examples
when solving the current task.
\end{lstlisting}
\caption{System Prompt for WorkBench}
\label{tab:workbench_system_prompt}
\end{figure*}

\begin{figure*}[t]
\begin{lstlisting}[basicstyle=\ttfamily\small, frame=single, escapeinside={(*}{*)}]
1) Delete my last email from nadia (*$\rightarrow$*) Remove the most recent email I received from
Nadia

2) All my emails from yuki from the last 3 days need to be deleted. Can you do that?
(*$\rightarrow$*) Remove Yuki's emails that arrived in the previous three days

3) carlos needs all the emails from chenwei last week about 'Update on Supply Chain
Enhancement Workshop'. Can you forward them? (*$\rightarrow$*) Could you forward to Carlos every
email Chenwei sent last week about 'Update on Supply Chain Enhancement Workshop'?
\end{lstlisting}
\caption{Examples of original WorkBench email queries and their paraphrased forms}
\label{tab:workbench_paraphrased_queries}
\end{figure*}

\begin{figure*}[t]
\begin{lstlisting}[basicstyle=\ttfamily\small, frame=single, escapeinside={(*}{*)}]
You are an expert system designed to generate realistic queries that are executable
using available tools to create high-quality training data for tool-use agents.

## Your Objectives:
Generate diverse, realistic user queries that can be solved using the available
tools

## Available Tools:
{TOOLS}

## Generation Guidelines:

Today's date is Thursday, 2023-11-30 and the current time is 00:00:00.
Remember the current date and time when answering queries. Meetings must not start
before 9am or end after 6pm.

### Query Diversity:
- Include multi-tool scenarios
- Create queries that require different reasoning patterns (sequential, parallel,
conditional)
### Query Quality Standards:
- Queries should be realistic and represent actual user needs
- Use natural language that real users would employ
- Avoid overly artificial or contrived scenarios
### Execution Requirements:
- Follow proper tool calling conventions and error handling
- Provide clear reasoning for tool selection and parameter choices
- Each query should include at least one write operation
- Queries should not give explicit information

\end{lstlisting}
\caption{System prompt for the synthetic WorkBench query generation}
\label{tab:workbench_synthetic_query_prompt}
\end{figure*}

\begin{figure*}[t]
\begin{lstlisting}[basicstyle=\ttfamily\small, frame=single, escapeinside={(*}{*)}]
1) Find emails from Lena about homepage alignment issues from October, delete them
since that project is done, and send her an email about the new homepage design
meeting scheduled for December 5th at 11am.

2) Forward Fatima the latest quarterly tech roundup email from Dmitri. I want to 
make sure she's up to speed before our meeting tomorrow.

3) I need to reply to Santiago's latest email about the UX feedback submission
improvements. Tell him I've reviewed it and it looks great, but I'd like to discuss
the project timeline adjustments he mentioned next week.

\end{lstlisting}
\caption{Example synthetic queries for WorkBench email domain}
\label{tab:workbench_synthetic_query_email}
\end{figure*}

\begin{figure*}[t]
\begin{lstlisting}[basicstyle=\ttfamily\small, frame=single, escapeinside={(*}{*)}]
Your task is to use a tool that not only meets real-time user instructions but also
aligns with user preferences.

These preferences can be inferred from the user's interaction history provided 
below, including past user instructions and corresponding tool calls, with user's 
binary ratings reflecting the user's satisfaction with the tool calls.

Interaction history is:

instruction 1: I want to search for the TV show 'Stranger Things' with
an offset of 0 and limit suggestions to 20 and limit titles to 50

tool_call 1:
{"tool_name": "<Data>.<Netflix Data>.<search>, "parameters": {"query":
"stranger", "offset": 0, "limit_suggestions": 20, "limit_titles": 50}},
"rating": 1},

instruction 2: .....
tool_call 2: ....
.
.
.

instruction N: .....
tool_call N: ....

Available tools you can call with input parameters are listed here:

{TOOLS}

Remember:
Generate only the tool call in JSON format without any additional outputs.

You will now receive a user instruction; please generate the tool call. 

Begin! 

\end{lstlisting}
\caption{Original system prompt for PEToolBench}
\label{tab:petoolbench_original_prompt}
\end{figure*}

\begin{figure*}[t]
\begin{lstlisting}[basicstyle=\ttfamily\small, frame=single, escapeinside={(*}{*)}]

You are a preference analysis assistant that examines tool usage patterns across
queries to identify user preferences based on binary ratings (1 = positive, 0 =
negative). Your task is to analyze the current query and historical queries with
their tool executions and binary ratings.

Current query, its execution and binary rating is:
{current_execution}

Historical queries, their executions and binary ratings:
{retrieved_memories}

Rules:
IMPORTANT: Only consider it a preference pattern if a domain, API or tool appears in
at least TWO queries with positive ratings (1) or negative ratings (0))
- Only report preferences where there is a clear pattern
- If no clear preferences exist at any level, state "No clear preference identified"
- Only consider the preferences related to the current query

Output your analysis in the following format:
***PREFERENCES***:
- Domain Preferences: [List domains that show consistent positive or negative
ratings]
- API Preferences: [List APIs that show consistent positive or negative ratings]

Example Output
***PREFERENCES***:
- Domain Preferences: <Sports>

- API Preferences: <Sports>.<Football Data>
- Domain Nonpreferences (if there is any): <Health>
- API Nonpreferences (if there is any): <Health>.<Heart Rhythm Data>

\end{lstlisting}
\caption{Memory extraction module's system prompt for PEToolBench}
\label{tab:petoolbench_thinker_prompt}
\end{figure*}

\begin{figure*}[t]
\begin{lstlisting}[basicstyle=\ttfamily\small, frame=single, escapeinside={(*}{*)}]
Domain Preferences: <Business>
API Preferences: <Business>.<Idealista_v2>, <Business>.<Bayut>

\end{lstlisting}
\caption{Example memory extractor output for PEToolBench}
\label{tab:petoolbench_thinker_output}
\end{figure*}

\begin{figure*}[t]
\begin{lstlisting}[basicstyle=\ttfamily\small, frame=single, escapeinside={(*}{*)}]

Example Query: A piece of work can finish by a certain number of men in 100 days.
If however, there were 10 men less, it would take 10 days more for the work to be
finished. How many men were there originally?

Tool Calls Used:
[{"name": "multiply", "label": "$var_1", "arguments": {"arg_0": 100, "arg_1": "x"}},
{"name": "subtract", "label": "$var_2", "arguments": {"arg_0": "x", "arg_1": 10}},
{"name": "multiply", "label": "$var_3", "arguments": {"arg_0": "$var_2.result$",
"arg_1": 110}},
{"name": "divide", "label": "$var_4", "arguments": {"arg_0": "$var_1.result$",
"arg_1": 100}},
{"name": "divide", "label": "$var_5", "arguments": {"arg_0": "$var_3.result$",
"arg_1": 110}},
{"name": "subtract", "label": "$var_6", "arguments": {"arg_0": "$var_5.result$",
"arg_1": "$var_4.result$"}},
{"name": "divide", "label": "$var_7", "arguments": {"arg_0": 1000, "arg_1": 10}},
{"name": "divide", "label": "$var_8", "arguments": {"arg_0": "$var_7.result$",
"arg_1": 10}}]

Reflection/Critic: 

The agent's approach to solve this work problem contains multiple errors. Let's
examine the problem and the agent's solution:

Problem Understanding
We have a certain number of men (let's call it x) who can finish work in 100 days
With 10 fewer men (x-10), the same work takes 110 days (100+10 days)
Correct Approach
The key insight is that work = men (*$\times$*) days, and the total work is constant. So:
(*$x \times 100 = (x-10) \times 110$*)
Errors in the Agent's Solution:
Parameter Type Error: In the first function call, the agent uses "x" as a string
instead of a numeric variable, which would cause execution issues.
Mathematical Logic Error: The agent's sequence of operations doesn't correctly set
up or solve the equation. The proper approach requires solving the equation (*$x \times 100 = (x-10) \times 110$*).
Unnecessary Operations: The last two divisions with 1000 and 10 don't relate to the
problem's solution and appear arbitrary.
Missing Final Calculation: The agent doesn't actually solve for the original number
of men.
Correct Solution Would Be:
Set up the equation: (*$x \times 100 = (x-10) \times 110$*)
Expand: (*$100x = 110x - 1100$*)
Rearrange: (*$100x - 110x = -1100$*)
Simplify: (*$-10x = -1100$*)
Solve: (*$x = 110$*)
The original number of men was 110.
The agent failed to properly formulate and solve the algebraic equation needed for
this problem.

\end{lstlisting}
\caption{Example memory entry for NESTFUL}
\label{tab:nestful_memory_entry}
\end{figure*}

\begin{figure*}[t]
\begin{lstlisting}[basicstyle=\ttfamily\scriptsize, frame=single, escapeinside={(*}{*)}]


Query: Two employees X and Y are paid a total of Rs. 440 per week by their
employer. If X is paid 120 percent of the sum paid to Y, how much is Y paid per 
week?

You are a helpful assistant with access to the following function calls.
Your task is to produce a sequence of function calls necessary to generate response
to the user utterance. Here is a list of functions in JSON format that you can 
invoke: {TOOLS}

DO NOT try to answer the user question, just invoke the tools needed to respond to
the user, if any. The output MUST strictly adhere to the following JSON format:
[{"name": "func_name1", "arguments": {"argument1": "value1", "argument2": "value2"},
"label": "$var_1"}, ... (more tool calls as required)]. Please make sure
the parameter type is correct and follow the documentation for parameter format.
If no function call is needed, please directly output an empty list.

Here are some examples and what was wrong with the agent response:
Example Query: Two employees X and Y are paid a total of Rs. 506 per week by their
employer. If X is paid 120 percent of the sum paid to Y, how much is Y paid per 
week?
Tool Calls Used:
[{"name": "add", "label": "$var_1", "arguments": {"arg_0": 1, "arg_1": 1.2}},
{"name": "divide", "label": "$var_2", "arguments": {"arg_0": 506, "arg_1":
"$var_1.result$"}},
{"name": "multiply", "label": "$var_3", "arguments": {"arg_0": "$var_2.result$",
"arg_1": 1}}]
Reflection/Critic: correct tool calls
========================================
Example Query: Rs.385 were divided among X,Y,Z in such a way that X had Rs.20 more
than Y and Z had Rs 15 more than X. How much was (*$Y$*)'s share?
Tool Calls Used:
[{"name": "add", "label": "$var_1", "arguments": {"arg_0": 20, "arg_1": 15}},
{"name": "subtract", "label": "$var_2", "arguments": {"arg_0": 385, "arg_1":
"$var_1.result$"}},
{"name": "divide", "label": "$var_3", "arguments": {"arg_0": "$var_2.result$",
"arg_1": 3}}]
Reflection/Critic: I need to analyze the incorrect tool execution for solving this
problem about dividing Rs.385 among X, Y, and Z with specific conditions.

The mistake in the agent's execution:
The agent tried to solve the problem by:
1. Adding 20 + 15 = 35
2. Subtracting this from 385 to get 350
3. Dividing 350 by 3
This approach is incorrect because it misunderstands the relationships between X,
Y, and Z.
The correct approach should account for:
- X has Rs.20 more than Y (so X = Y + 20)
- Z has Rs.15 more than X (so Z = X + 15 = Y + 20 + 15 = Y + 35)
If we call Y's share Y, then:
- X = Y + 20
- Z = Y + 35
The total is 385:
Y + (Y + 20) + (Y + 35) = 385
3Y + 55 = 385
3Y = 330
Y = 110
The agent should have:
1. Added 20 + 15 = 35 (for the total of additional money beyond Y)
2. Added 35 to Y to represent Z (not needed as a separate step)
3. Realized that the equation is 3Y + 55 = 385
4. Subtracted 55 from 385 to get 330
5. Divided 330 by 3 to get Y = 110
The mistake was in not properly setting up the equation to account for all three
shares in terms of Y before solving.

Use these insights to avoid similar mistakes. Consider these examples when
solving the current task.

\end{lstlisting}
\caption{Example NESTFUL system prompt with a test query and retrieved memory entries}
\label{tab:nestful_test_query_and_memories}
\end{figure*}

\end{document}